\documentclass[conference]{IEEEtran}
\IEEEoverridecommandlockouts
\usepackage{cite}
\usepackage{amsmath,amssymb,amsfonts}
\usepackage{algorithmic}
\usepackage{graphicx}
\usepackage{svg}
\usepackage{multirow}
\usepackage{multicol}
\usepackage{xcolor}
\def\BibTeX{{\rm B\kern-.05em{\sc i\kern-.025em b}\kern-.08em
    T\kern-.1667em\lower.7ex\hbox{E}\kern-.125emX}}

\usepackage{booktabs}
\usepackage{svg}
\usepackage{slashbox}

\usepackage[breaklinks, colorlinks]{hyperref}

\begin{document}

\title{BASED: Benchmarking, Analysis, and Structural Estimation of Deblurring
\thanks{This study was supported by Russian Science Foundation under grant 22-21-00478, \href{https://rscf.ru/en/project/22-21-00478/}{https://rscf.ru/en/project/22-21-00478/}}
}

\author{\IEEEauthorblockN{Nikita Alutis\IEEEauthorrefmark{1}\quad Egor Chistov\IEEEauthorrefmark{1}\quad Mikhail Dremin\IEEEauthorrefmark{1}\quad Dmitriy Vatolin\IEEEauthorrefmark{1}\IEEEauthorrefmark{2}}
\IEEEauthorblockA{\textit{Lomonosov Moscow State University\IEEEauthorrefmark{1}} \\
\textit{MSU Institute for Artificial Intelligence\IEEEauthorrefmark{2}}\\
Moscow, Russia \\
\texttt{ \{nikita.alutis, egor.chistov, mikhail.dremin, dmitriy\}@graphics.cs.msu.ru}}
}

\maketitle
\begin{abstract}
This paper discusses the challenges of evaluating deblurring-methods quality and proposes a reduced-reference metric based on machine learning. Traditional quality-assessment metrics such as PSNR and SSIM are common for this task, but not only do they correlate poorly with subjective assessments, they also require ground-truth (GT) frames, which can be difficult to obtain in the case of deblurring. To develop and evaluate our metric, we created a new motion-blur dataset using a beam splitter. The setup captured various motion types using a static camera, as most scenes in existing datasets include blur due to camera motion. We also conducted two large subjective comparisons to aid in metric development. Our resulting metric requires no GT frames, and it correlates well with subjective human perception of blur.
\end{abstract}


\section{Introduction}
Numerous learning-based approaches to both single-image and video deblurring have recently been proposed. Today, however, there is no comprehensive way to assess deblurring quality. Instead, common methods for this task include the traditional PSNR and SSIM\cite{b1} image quality metrics, as well as the deep-learning-based LPIPS metric \cite{b2} and super-resolution quality ERQA metric \cite{b3}. However as we show later, they all correlate poorly with subjective human perception of blur, and they require GT-frames, which can be hard to obtain when creating a deblurring dataset. 

The best way to assess methods performance is to conduct a subjective comparisons, but they are too costly to employ every time a method needs evaluation. Therefore, we propose a new machine-learning-based reduced-reference metric to specifically evaluate deblurring quality.

 Training the metric and evaluating its performance requires datasets with real blur, because as \cite{b6} proves, real blur differs considerably from synthetic blur. Three recent datasets created using a beam-splitter have emerged \cite{b4, b5, b6}; among them, RSBlur \cite{b6} is the most diverse.

 \begin{figure}[htb]

\begin{minipage}[b]{1.0\linewidth}
  \centering
  \centerline{\includegraphics[width=8.5cm]{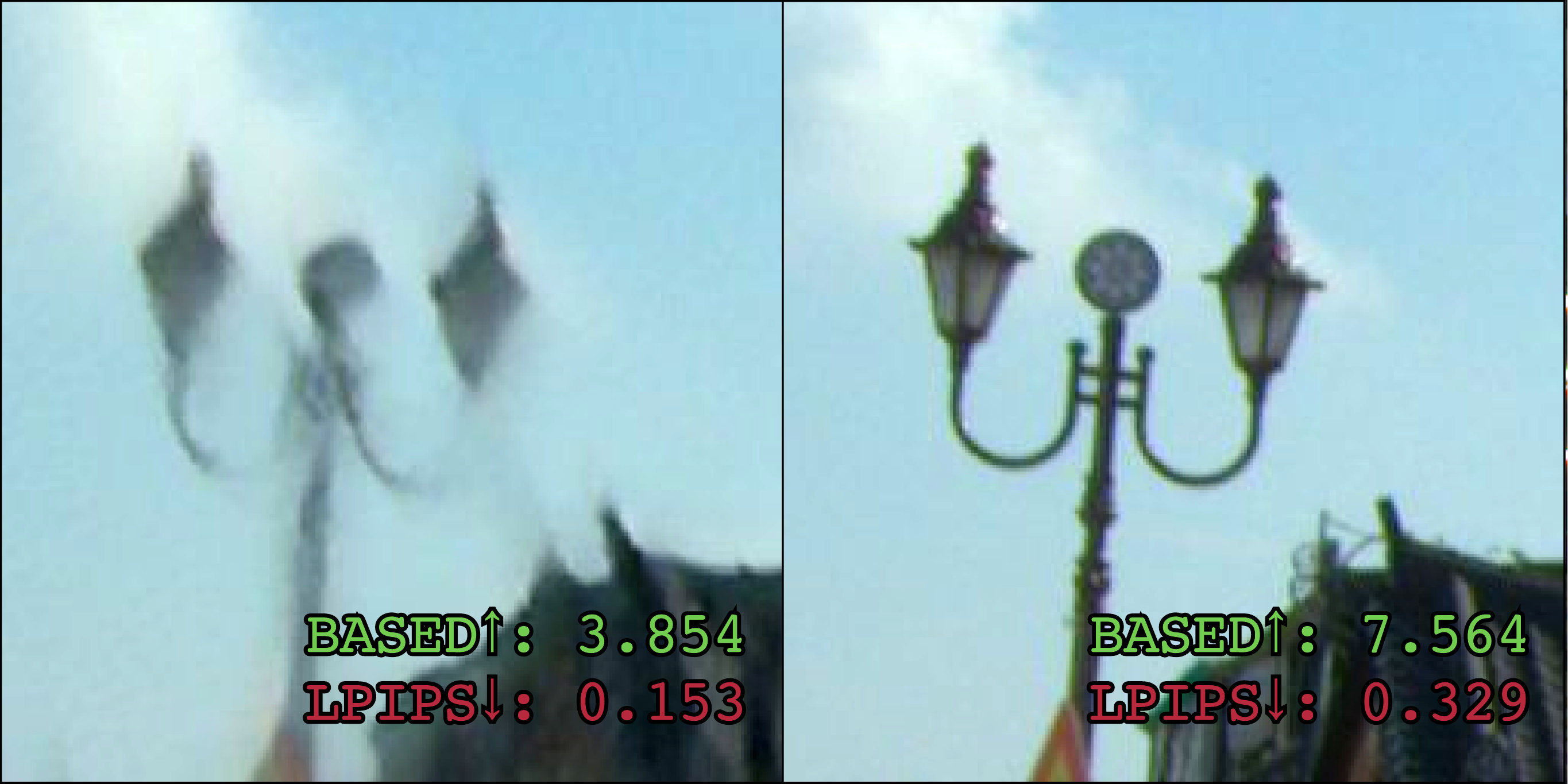}}
\end{minipage}

  \caption{Example of BASED and LPIPS metrics on crop from RSBlur dataset. Right image is significantly less blurred, however LPIPS shows that right image is more blurred than left, while BASED metric accurately reflects increase in sharpness}
  \label{fig:comparison}

\end{figure}

 Even this dataset has a drawback, however: it mostly contains scenes in which blur results from camera motion, so the blurring is nearly even over the entire frame.  But in the real world, blur often occurs due to the movement of one or more fast-moving objects in the frame. These videos are much more difficult for neural networks to correctly process (as we show later), because only a small part of the image actually suffers from blurring; other regions provide little or no information to aid in restoration. 
 
 Therefore, to ensure maximum generalization ability of the resulting metric, we created the Benchmarking, Analysis, and Structural Estimation of Deblurring (BASED) dataset using a beam-splitter setup that employs a static camera to capture a variety of motion types.
 
 Finally, to train the metric to correlate with human perception, we needed to get subjective values for each crop of the dataset. Thus, we performed two large independent subjective comparisons on both the BASED and RSBlur datasets, each using a slightly different sets of methods.
 
 Our contributions are the following:
\begin{itemize}
\item We created a real-motion-blur dataset using a beam-splitter setup and tested the performance of state-of-the-art deblurring methods.
\item We developed a comprehensive deblurring benchmark to quantify the ability of deblurring models to restore blurred videos. The benchmark is publicly available at \href{https://videoprocessing.ai/benchmarks/deblurring.html}{https://videoprocessing.ai/benchmarks/deblurring.html}.
\item We conducted two large subjective comparisons to analyze existing quality metrics and collected  reference values to aid in developing our metric.
\item We produced a machine-learning-based metric that evaluates deblurring quality without requiring GT frames.
\end{itemize}

The code for the resulting metric can be found at \href{https://github.com/illaitar/based}{https://github.com/illaitar/based}.

\section{Related Works}
Many neural-network-based deblurring methods are in use today \cite{b7, b8, b9, b10, b11, b12, b13, b15, b14}. Some are image based, including MFFDNet\cite{b7} and NAFNet\cite{b8}, and some are video based, including VRT\cite{b10}. Most research articles that propose such methods employ traditional full-reference image quality assessment through, for example, PSNR or SSIM. The ERQA metric for super-resolution-quality assessment has also served to assess deblurring quality. And recently, several neural-network-based metrics have emerged, the most popular of which is LPIPS, but they rarely appear in such articles. Researchers have also created specialized no-reference blur metrics, such as \cite{b18} and \cite{b21}. Our aim is to evaluate how well all these metrics handle image and video deblurring.

Datasets for deblurring typically fall into two categories: synthetic blur, such as GoPro \cite{b16} and Reds \cite{b17}, and real-world blur, such as BSD\cite{b4}, RealBlur\cite{b5}, and RSBlur\cite{b6}. Synthetic blur, as \cite{b6} shows, differs too greatly from real-world blur to judge a method’s performance on arbitrary videos. At the same time, real-blur datasets all suffer from problems related to image and data processing:\begin{itemize}
    \item BSD, RealBlur, and RSBlur contain only images, not videos.
    \item BSD has low resolution.
    \item RealBlur mostly includes low-light scenes, but real-world blur often occurs in well-lit scenes.
    \item All three datasets are mostly based on camera motion, so the frames exhibit even blurring. In the real world, however, blur is often due to the motion of one or more fast-moving objects; therefore, only some areas in these videos contain this artifact.
\end{itemize}

\section{Proposed Dataset}
We propose a new dataset of videos containing real-world motion blur. We collected it using a beam-splitter optical device.

\subsection{Scene Selection}

We filmed various objects and motion types to diversify the scenes. Motion types in our dataset include swinging, falling, rolling, linear movement, and turning. We set up a white background and a static light source to focus on the blurred objects. Out of 135 videos we handpicked 23 to represent movement type and speed, texture complexity, and temporal and spatial complexity.

\subsection{Recording Rig}
Our beam-splitter rig consisted of the beam-splitting glass, which has a 0.5 reflection coefficient, and two cameras with zero visual stereo base between them. We used a setup similar to that used for the RealBlur dataset.

A box protected our cameras and the beam splitter from external light; its inside was covered with matte light-absorbing fabric. This arrangement eliminated the camera-lens reflections that previous datasets exhibited. We disabled optical stabilization and manually fixed all settings, such as ISO, shutter speed, and color temperature.

To create the dataset, we had to simultaneously capture blurred (target) and nonblurred (reference) videos. To do so, our two cameras employed different exposure times. We set the one filming the blurred view to a 0.1-second exposure and the one filming ground truth to 0.005 seconds. The other settings were the same for both cameras, so all splitter-glass distortions were insignificant. The color-correction step compensated for brightness variation due to different shutter speeds.

\subsection{Parallax Minimization}
To obtain accurate and reliable ground-truth data for a deblurring dataset, filming scenes with minimal parallax is essential.  Parallax refers to the apparent shift in an object's position when viewed from different lines of sight. Although post-processing techniques can effectively correct for affine mismatches between cameras, such as scale, rotation, and translation, correcting for parallax is more difficult. Eliminating parallax at this stage requires the use of complex optical flow and inpainting algorithms, which can potentially compromise the quality of the ground-truth data.

Parallax correction is crucial for several reasons. First, it helps maintain the dataset’s integrity  by ensuring that the captured images are consistent and reliable. This, in turn, enables the development of more accurate and robust deblurring algorithms. Second, minimizing parallax simplifies the data processing pipeline by reducing the need for complex and computationally expensive optical-flow algorithms. This approach streamlines the development process and saves valuable time and resources.

Parallax minimization can employ a two-step process. First, the lenses of the left distorted and left ground-truth cameras must be visually aligned to create a visual zero stereo base. This alignment ensures that the cameras share a common point of view, reducing the potential for parallax. Next, the cameras must aim at the same scene to maximize the overlapping region, using a photo captured by one camera and a video stream captured by the other. Doing so enables direct comparison of the images to ensure no parallax is present.

Once the absence or near absence of parallax is confirmed, all scenes can be filmed in a single uninterrupted session. The ground-truth data is therefore consistent, reliable, and free of parallax-induced distortions --- vital features for the development of effective deblurring algorithms.

\subsection{Post-processing}
\begin{figure}[htb]

\centering
\includegraphics[width=1.0\linewidth]{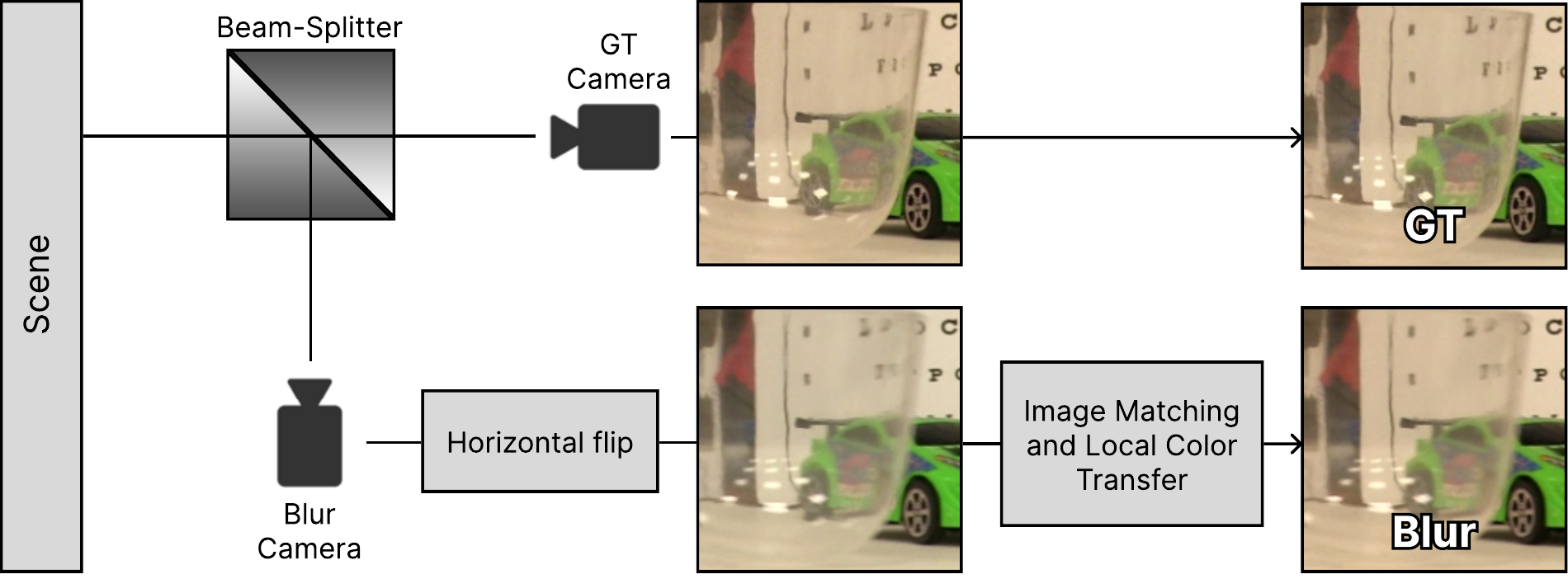}
\caption{Dataset-creation pipeline. Incoming light enters a beam splitter, and two cameras receive the same image. After initial filming we applied temporal, spatial, and color alignment.}
\label{fig:pipeline}
\end{figure}

Because the resulting views have little or no parallax present, we were able to implement a simple post-processing pipeline that avoids any significant frame distortion. We first flipped the target camera view horizontally, then we matched it to the ground truth view using a homography transformation. For the estimation of transformation parameters, we utilized SIFT\cite{b25} extractor, Brute-Force matcher, and the MAGSAC++ algorithm\cite{b24}. Global brightness difference and small color mismatches between two views were corrected using the method \cite{b22}. Temporal alignment employed audio streams captured by the cameras. Fig~\ref{fig:pipeline} presents a scheme of our pipeline.


\section{Benchmark}
We tested 7 methods and in total 11 their variations (i.e. trained on different datasets) on BASED: NAFNet (REDS), MAXIM (REDS), Restormer (GoPro), VRT (REDS),   MPR local (GoPro) , VRT (GoPro),  DeepRFT (GoPro),   DeepRFT (REDS),  MAXIM (GoPro), DeblurGAN Inception (GoPro),  Restormer local (GoPro). Our examination employed PSNR, SSIM, LPIPS, and ERQA to test the results. We selected these metrics because they are the most common in articles that cover deblurring. The comparison results are shown on Table~\ref{tab:metric}. Examples of methods work are present on Fig~\ref{fig:based}.

It is obvious that all metrics show a different picture in terms of methods performance. It can also be seen that clearly, some metrics provide misinformation in terms of methods performance. 

\section{Subjective Comparison}

\begin{figure*}
\includegraphics[width=1.0\linewidth]{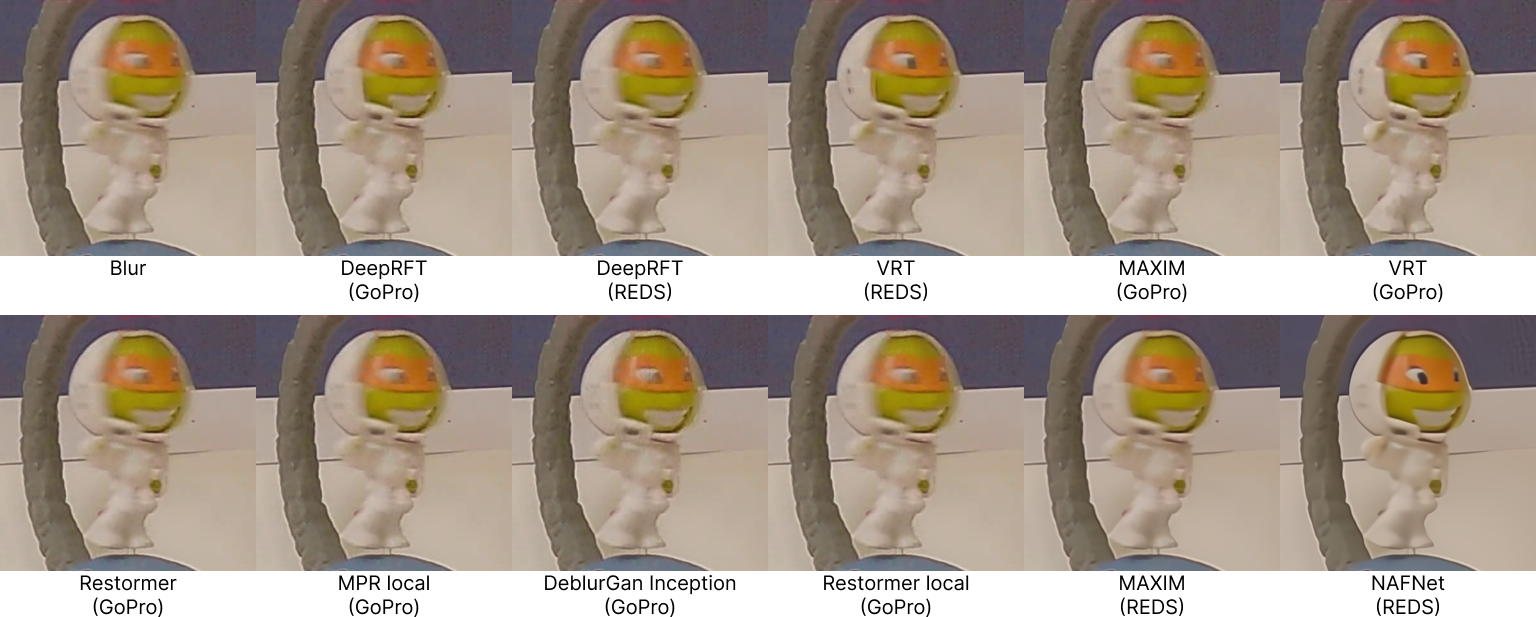}
\caption{Method results on BASED. NAFNet clearly shows the best result.}
\label{fig:based}
\end{figure*}

\begin{figure*}
\includegraphics[width=1.0\linewidth]{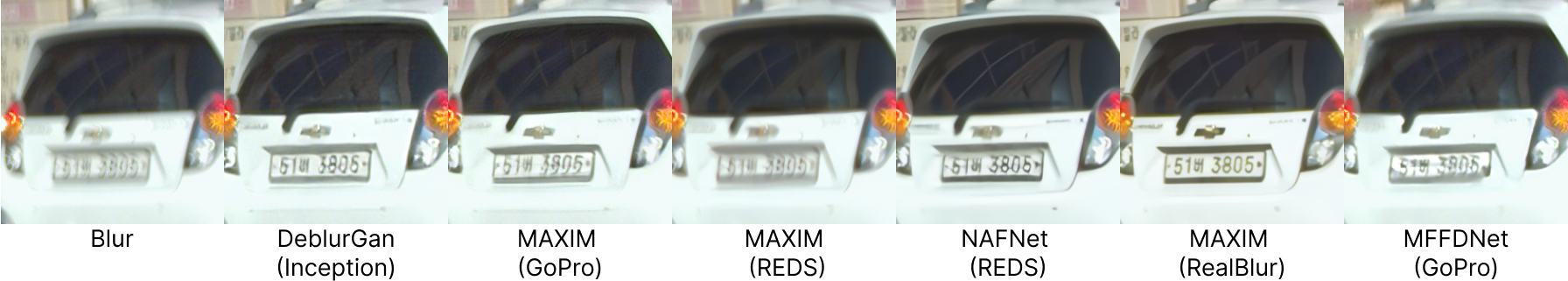}
\caption{Methods results on RSBlur. On this challenging example MAXIM, which was trained on real blur dataset, shows a better performance.}
\label{fig:rsblur}
\end{figure*}

We measured the performance of various groups of deblurring methods on the BASED and RSBlur datasets to obtain the most diverse dataset for metric training. We cropped the original dataset frames to three resolutions to test method performance at different scales: 512x512, 256x256, and 128x128. Our comparison used 253 crops from BASED and 1,200 from RSBlur. Crops from RSBlur and BASED are shown on Fig~\ref{fig:rsblur} and Fig~\ref{fig:based}.

We conducted a side-by-side pairwise evaluation using the Subjectify.us service, which enables crowdsourced subjective comparisons. To estimate information fidelity, we asked participants to avoid choosing the most beautiful video and instead choose the one that exhibits better detail restoration. Because participants are not experts in this field, they lack professional biases. Every participant was shown 25 video pairs and in each case was asked to choose the best video (“indistinguishable” was also an option). Each pair was shown to 10–15 participants until the confidence interval stopped changing. Of the 25 pairs for each participant, 3 are for verification, so the final results exclude those responses. Verification questions are necessary because we want to ensure that participants understand the task and avoid making random choices --- a critical facet of effective large-scale crowdsourced comparisons. All other responses from 1,904 successful RSBlur participants and 648 successful BASED participants went into calculating subjective scores using the Bradley-Terry model\cite{b19}.

\begin{figure}[htb]

\begin{minipage}[b]{1.0\linewidth}
  \centering
  \centerline{\includegraphics[width=8.5cm]{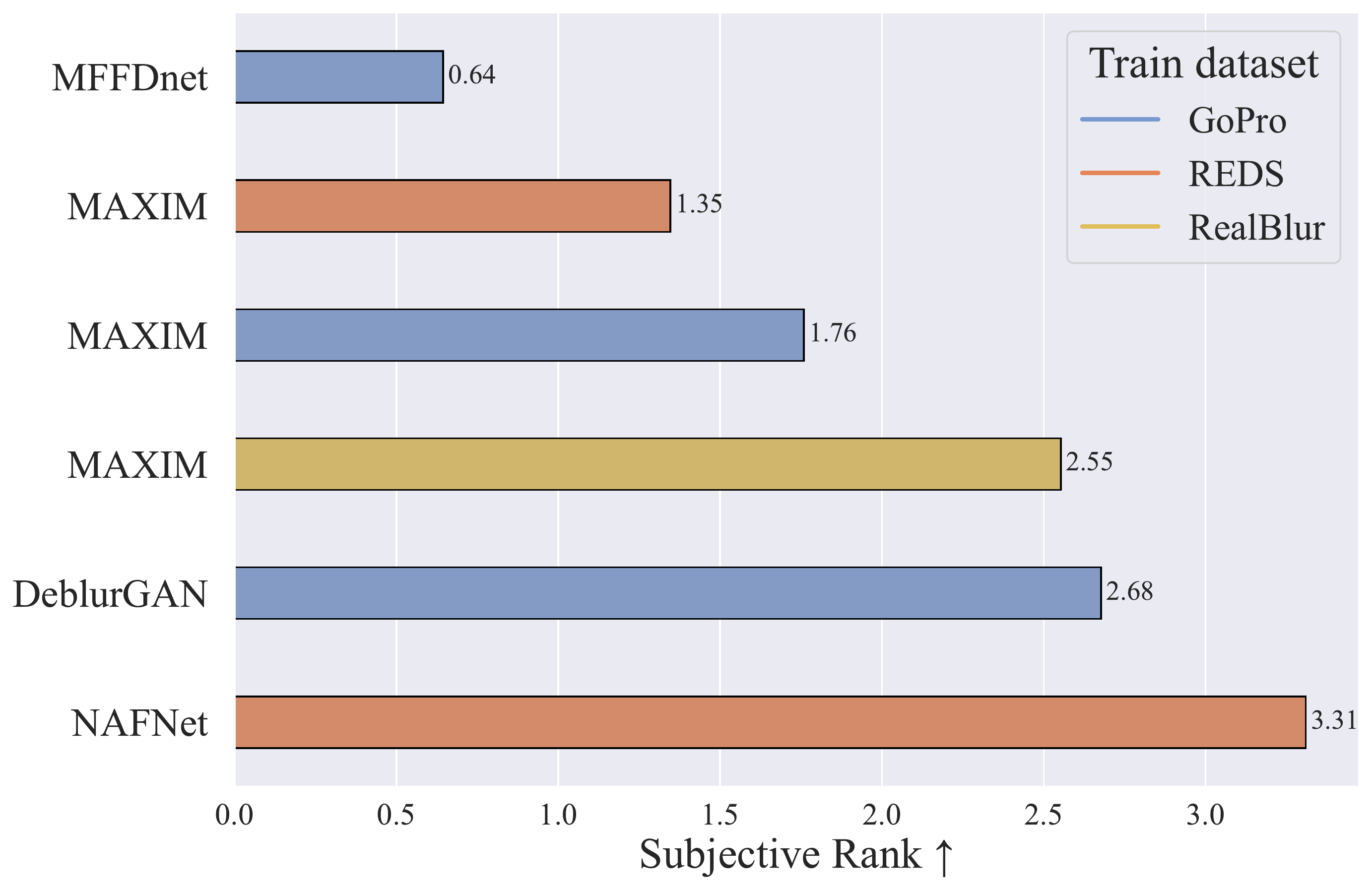}}
\end{minipage}

  \caption{Subjective-comparison results for RSBlur. The study comprises input from 1,904 participants. NAFNet performed much better than other methods.}
  \label{fig:subj}

\end{figure}

 Table~\ref{tab:metric} and Fig~\ref{fig:subj} present the results of our subjective comparison. NAFNet was the best method across two datasets.

\begin{table}
    \caption{Results using BASED. (G) indicates methods trained on the GoPro dataset, (R) indicates methods trained on the REDS dataset, and {L} indicates the local version of the method\cite{b12}. The best result appears in \textbf{bold}.}
  \centering
  \begin{tabular}{l|ccccc}
  \toprule
    \backslashbox{Method}{Metric} & PSNR$\uparrow$ & SSIM$\uparrow$ & LPIPS$\downarrow$ & ERQA$\uparrow$\ & Subj.$\uparrow$\ \\
  \midrule
    NAFNet (R) & 30.5480 & \textbf{0.9504} & 0.0856 & 0.7451 & \textbf{2.84}\\
    VRT (G) & 31.4295 & 0.9450 & 0.0817 & 0.7487 & 2.39\\
    VRT (R) & 30.9788 & 0.9460 & 0.0824 & \textbf{0.7506} & 1.57\\
    DeblurGAN (G) & 31.1717 & 0.9430 & 0.0887 & 0.7430 & 1.04\\
    MAXIM (R) & 30.6573 & 0.9496 & \textbf{0.0784} & 0.7428 & 1.01\\
    DeepRFT (G) & 31.5761 & 0.9448 & 0.0832 & 0.7432 & 0.54\\
    DeepRFT (R) & 31.3234 & 0.9448 & 0.0814 & 0.7434 & 0.46\\
    MPR \{L\} (G) & 31.6504 & 0.9454 & 0.0832 & 0.7452 & 0.44\\
    MAXIM (G) & 31.3634 & 0.9439 & 0.0919 & 0.7444 & 0.21\\
    Restormer (G) & \textbf{31.7611} & 0.9463 & 0.0824 & 0.7478 & 0.12\\
    Restormer \{L\} (G) & 31.1234 & 0.9421 & 0.0825 & 0.7387 & 0.12\\
  \bottomrule
  \end{tabular}
  \label{tab:metric}
\end{table}





\section{Proposed Metric}

\subsection{Concept}
We set out to create a reduced-reference metrict hat operates without GT frames. It takes two pictures --- an original blurred one and a resulting deblurred one, and provides a quality score.

We trained Random Forest Regressor \cite{b20} on several metrics traditionally used for texture analysis and sharpness estimation tasks. We tested the model with different hyper-parameters and found out that 220 estimators (Table~\ref{tab:estimators}) and a squared error criterion are the best combination. Our approach used the same crops as the subjective comparison, allowing us to employ the subjective values when training and testing the metric. We have also tried gradient boosting approach\cite{b26}, but it tended to overfit on both datasets.

We wanted to completely avoid overfitting on our data, so we trained the metric only on a smaller BASED dataset. Later, we tested it on RSBlur to prove its generalizability. In addition, our analysis did not use RSBlur correlations to tweak individual features; it only used BASED values as a reference.

\subsection{Features}
We studied and tweaked many features and these showed the best overall results in terms of individual correlation and correlation of whole metric on BASED:
\subsubsection{Laplacian}
The Laplacian is a second-order-derivative operator that enables detection of edges and intensity changes in an image. Assessing blur using this operator involves computing the Laplacian of an image and examining its properties to determine the degree of blur. 

\subsubsection{FFT}
The FFT is a mathematical technique that converts a signal from the time domain into the frequency domain, allowing us to analyze its spectral content. Assessing blur using the Fast Fourier Transform (FFT) involves analyzing the frequency content of an image in the frequency domain. 

For our task, we filtered all low frequencies with the standard threshold of 30, and then compared two resulting functions.

\subsubsection{Gabore Filtering}

Gabor filters are a type of linear filter used in image processing and computer vision to extract features from images. They are constructed by multiplying a Gaussian kernel and a complex sinusoidal wave, yielding a filter that is sensitive to both spatial location and frequency. The response of a Gabor filter to an image is obtainable by convolving the filter with the image. Doing so produces a filtered image that highlights areas of the original that match the filter’s frequency and orientation selectivity.

We used a set of three Gabore kernels to estimate their response and then calculated the difference between the resulting vectors.

\subsubsection{Hough}

The basic idea behind the Hough transform is to represent the image in a parameter space such that each point in the image corresponds to a curve in that space. 

\subsubsection{HOG}

Histogram of oriented gradients (HOG) is a feature-extraction technique by which computer-vision and image-processing algorithms detect object shapes and their boundaries. It divides an image into small cells, computes the gradient orientations and magnitudes for every pixel in each cell, and then aggregates the gradients into a histogram of gradient orientations for each cell.

To obtain the result of this metric, we calculated the L2 norm between the HOG for the original image and the HOG for the deblurred one.

\subsubsection{SSIM-M}
SSIM is a traditional metric used to measure the similarity between two images based on their structural content. We modified SSIM to work with YUV images and calculated the result as $6\times SSIM(Y) + SSIM(U) + SSIM(V)$, which has been proven to correlate better with human perception\cite{b23}.

\subsubsection{Sobel}
The Sobel operator works by convolving an image with two filters to calculate the image gradient. One filter detects horizontal edges, while the other detects vertical edges. After applying the filters, the resulting gradient images are combined to produce an edge map that highlights the edges in the original image.

We used Sobel kernels with the size of 13 and then calculated the L2 norm of the residual of two resulting gradient images.

\subsubsection{LBP}
Local binary patterns (LBP) is a histogram-based approach that represents an image’s local structure by computing a binary code for each pixel on the basis of its relationship with neighboring pixels.

The idea behind LBP is to compare a pixel’s intensity with those of surrounding pixels and generate a binary code for that pixel. Once it computes these codes for all pixels in the image, it generates a histogram that counts the frequency of each code.

To obtain a result for this metric, we calculated the L2 norm of two LBP histograms: one from the original image and one from the deblurred image.

\subsubsection{Reblur}
We used the approach from \cite{b21} to calculate the difference between the reblurred original image and reblurred blur image using the L2 norm of their absolute difference. Our version applied a Gaussian kernel of size of 17 to the reblur images.

\begin{table}
    \caption{Mean Pearson (PLCC), Spearman (SRCC), and Kendall (KRCC) correlation coefficients between metric features and subjective-comparison results. The proposed metric was trained on BASED dataset. For results on BASED, 5 fold cross validation was used as final score. The best result appears in \textbf{bold}.}
  \centering
  \begin{tabular}{l|cccccc}
  \toprule
           & \multicolumn{3}{c}{\textbf{RSBlur}} & \multicolumn{3}{c}{\textbf{BASED}} \\
    Metric & PLCC & SRCC & KRCC & PLCC & SRCC & KRCC\\
  \midrule
    Laplacian & \textbf{0.8584} & 0.8343 & 0.7421 & 0.8939 & 0.8716 & 0.7521\\
    FFT & 0.7356 & 0.8411 & 0.7504 & 0.8327 & 0.8613 & 0.7430\\
    Reblur & 0.6950 & 0.6672 & 0.5593 & 0.6945 & 0.6432 & 0.5284 \\
    Hough & 0.8077 & 0.7918 & 0.6859 & 0.6718 & 0.7500 & 0.6259 \\
    Sobel & 0.7553 & 0.7652 & 0.6714  & 0.8984 & 0.8800 & 0.7476 \\
    HOG &  0.7188 & 0.7927 & 0.7007 & 0.8745 & 0.8595 & 0.7323  \\
    LBP & 0.6914 & 0.7830 & 0.6879 & 0.7653 & 0.8058 & 0.6713  \\
    Gabore & 0.5735 & 0.5965 & 0.5036 & 0.8831 & 0.8445 & 0.7233  \\
    SSIM-M & \textbf{0.8584} & 0.8188 & 0.7250 & 0.9242 & 0.8777 & 0.7521  \\
    \textbf{Proposed} & 0.8349 & \textbf{0.8520} & \textbf{0.7635} & \textbf{0.9531} & \textbf{0.9053} & \textbf{0.8821}  \\

  \bottomrule
  \end{tabular}
  \label{tab:comps}
\end{table}

\begin{table}
    \caption{Mean Pearson (PLCC), Spearman (SRCC) and Kendall (KRCC) correlation coefficients between metrics and subjective-comparison results.  The best result appears in \textbf{bold}.}
  \centering
  \begin{tabular}{l|cccccc}
  \toprule
           & \multicolumn{3}{c}{\textbf{RSBlur}} & \multicolumn{3}{c}{\textbf{BASED}} \\
    Metric & PLCC & SRCC & KRCC & PLCC & SRCC & KRCC\\
  \midrule
    SSIM & 0.8177 & 0.7623 & 0.6489 & 0.8821 & 0.8555 & 0.7216  \\ 
    PSNR & 0.7836 & 0.7644 & 0.6714 & 0.9044 & 0.8753 & 0.7476  \\
    LPIPS & 0.7520 & 0.7558 & 0.6290 & 0.9290 & 0.8833 & 0.7688  \\
    ERQA & 0.6980 & 0.7006 & 0.5983 & 0.8052 & 0.7741 & 0.6577  \\
    CPBD & 0.1176 & 0.0811 & 0.0618 & 0.0157 & 0.1112 & 0.1015 \\
    Reblur & 0.6747 & 0.6223 & 0.5132 & 0.6952 & 0.6408 & 0.5239 \\
    \textbf{Proposed} & \textbf{0.8349} & \textbf{0.8520} & \textbf{0.7635} & \textbf{0.9531} & \textbf{0.9053} & \textbf{0.8821}  \\
    
  \bottomrule
  \end{tabular}
  \label{tab:corrs}
\end{table}

\begin{table}[htb]
    \caption{Dependence of metric performance quality on the number of trees in the ensemble. The best result appears in \textbf{bold}.}
  \centering
  \begin{tabular}{l|ccc}
  \toprule
           & \multicolumn{3}{c}{\textbf{RSBlur}} \\
    Estimators & PLCC & SRCC & KRCC \\
  \midrule
    50 & 0.8312 & 0.8387 & 0.7434  \\ 
    100 & 0.8334 & 0.8512 & 0.7595  \\ 
    150 & 0.8323 & 0.8475 & 0.7546  \\ 
    200 & 0.8341 & 0.8502 & 0.7610  \\ 
    \textbf{220} & 0.8349 & \textbf{0.8520} & \textbf{0.7635} \\
    250 & 0.8349 & 0.8509 & 0.7617 \\ 
    300 & 0.8348 & 0.8487 & 0.7574 \\ 
    350 & 0.8352 & 0.8484 & 0.7581   \\ 
    400 &  \textbf{0.8353} & 0.8493 & 0.7586  \\ 
  \bottomrule
  \end{tabular}
  \label{tab:estimators}
\end{table}

\subsection{Results}
We achieved good results for all three correlations. The results of each individual component can be observed on Table~\ref{tab:comps}. Relative to traditional metrics, such as SSIM, PNSR, ERQA, and LPIPS, we also achieved state-of-the-art performance (Table~\ref{tab:corrs}). But these metrics were not designed for a reduced-reference analysis (i.e., without the use of GT frames) and thus exhibit worse results than they do in the full-reference case. We also tested openly available No-Reference blur metrics: the CPBD metric and the original version of Reblur\cite{b21} which we previously modified to use as one of the components.

\section{Conclusion}

In this paper we created a real-motion-blur dataset using a beam splitter and used it to test the performance of state-of-the-art deblurring methods. We also conducted two large subjective comparisons, one on our dataset and one on the existing RSBlur dataset, to both show the failure of current quality metrics and to collect reference values for metric development. Finally, we developed a machine-learning-based metric to evaluate deblurring quality without using GT frames or subjective comparisons.

\section{Acknowledgments}
Dataset preparation and subjective comparison were supported by Russian Science Foundation under grant 22-21-00478. Autors are also grateful to Daniil Davydov and Alexander Fominykh for detailed discussions of the study.


\end{document}